\documentclass[10pt,journal,compsoc]{IEEEtran}
\usepackage{xcolor}
\usepackage{amsmath, amsfonts}
\usepackage{algorithm}
\usepackage{algpseudocode}
\usepackage[utf8]{inputenc}
\usepackage{threeparttable}
\usepackage{algorithmicx} 
\usepackage{algpseudocode} % Defines \State, \Procedure etc.

\usepackage{amssymb} % For \mathbb{R}
\usepackage{bm} % For bold math symbols \bm
\usepackage{multirow} 
 \usepackage{graphicx}
 \usepackage{subcaption}
 \usepackage{booktabs}
 \usepackage{makecell}
 \usepackage{nicematrix}
 \usepackage[table]{xcolor}
\usepackage{hyperref} 
 
\ifCLASSOPTIONcompsoc
  \usepackage[nocompress]{cite}

\else
  \usepackage{cite}
\fi
\ifCLASSINFOpdf
\else
\fi
\begin{document}
\title{MIDiff: Tackling Sparsity and Imbalance in Mobile Usage Generation via Multivariate-Imaging Diffusion}

\author{Yilai~Liu, Shiyuan~Zhang, and~Hongyang~Du% <-this % stops a space
%后面加thank要把\protect \\ 加上
\IEEEcompsocitemizethanks{
\IEEEcompsocthanksitem Y. Liu, S. Zhang, and H. Du are with the Department of Electrical
and Computer Engineering, The University of Hong Kong, Pok Fu Lam,
Hong Kong SAR, China (e-mail: {shiyuanzhang, yilai\_liu}@connect.hku.hk; duhy@hku.hk).}% <-this % stops an unwanted space
}


\markboth{}{}
\IEEEtitleabstractindextext{%
\begin{abstract}
Mobile usage traces are critical for tasks such as user behavior prediction and app recommendation, yet their use is constrained by privacy restrictions and costly large-scale data collection. Although generative models perform well on general time series, their application to mobile usage data remains challenging because (i) limited user activity causes severe sparsity, (ii) heterogeneous variable types complicate joint modeling, and (iii) functional differences across apps create pronounced usage imbalance. To address these challenges, we propose Multivariate-Imaging Diffusion (MIDiff), a diffusion-based framework operating in an imaging space defined by Cross-Gramian Angular Sum Field (C-GASF). C-GASF transforms sparse multivariate sequences into correlation images, while MIDiff employs Triple Attention in a U-Net to preserve temporal consistency and variable dependencies. Experiments show that MIDiff achieves state-of-the-art performance across fidelity metrics. In particular, it obtains a Discriminative Accuracy (DA) of 0.1526, compared with 0.3476 for the strongest baseline, ZITS-VAE, demonstrating its effectiveness in generating realistic and diverse mobile usage traces. Our code is available at \href{https://github.com/YilaiLiu-HKU/MIDiff}{https://github.com/YilaiLiu-HKU/MIDiff}.
\end{abstract}

}


\maketitle
\IEEEdisplaynontitleabstractindextext

% \fi
\IEEEpeerreviewmaketitle

\IEEEraisesectionheading{\section{Introduction}\label{sec:introduction}}
% The rapid proliferation of mobile devices and applications is generating vast amounts of mobile usage traces.
% These traces contain rich information about complex mobile usage habits. 
% Such features enable operators and service providers to support a diverse array of applications, like bandwidth allocation and personalized app recommendation~\cite{lu2020user, ding2021no,apprec,nettra}. 
% These applications are essential for enhancing the end-user's Quality of Experience (QoE). 
The rapid proliferation of mobile devices and applications has generated vast amounts of mobile usage traces rich in complex behavioral information, enabling operators and service providers to support a diverse array of applications such as bandwidth allocation and personalized app recommendation~\cite{lu2020user, ding2021no, apprec, nettra}. 
These data-driven applications are essential for enhancing the end-user's Quality of Experience (QoE), yet their effectiveness hinges on access to large-scale, fine-grained usage data~\cite{yin2022practical, zuppelli2021pcapstego, jiang2023generative}.
However, due to privacy concerns and financial costs, the vast majority of such real-world traces remain inaccessible to most researchers and companies.
A practical alternative is to synthesize such data through generative models, as it can produce large-scale, diverse training data on demand without the constraints of real-world collection.

Existing generative models have achieved considerable success on general time series datasets~\cite{energy,TimeGANStock}. 
For instance, TimeGAN~\cite{TimeGANStock} enforces sequential dependencies via supervised loss, while TTS-GAN~\cite{TTSGAN} leverages self-attention to capture long-range correlations.
These methods share a common assumption that observations are dense and regularly sampled, allowing temporal dependencies to be reliably estimated. 
However, mobile usage traces fundamentally violate this assumption. User activity is intermittent and sparse, occurring only during brief, isolated sessions~\cite{sparse_1,sparse_2}, which makes stable temporal dependencies difficult to learn~\cite{lambert1992zero}. 
Another line of work, including Time-VAE~\cite{TimeVAE} and Diffusion-TS~\cite{yuan2024diffusionts}, attempts to model interpretable features such as periodicity and trends. Nevertheless, sparse mobile usage makes trend estimates unstable, and highly personalized behavior undermines interpretability of the periodic components.

\begin{figure*}[!t] % [!ht] 是一个可选参数，让LaTeX尽量把图片放在这里 (Here) 或顶部 (Top)
  \centering
\includegraphics[width=1\textwidth]{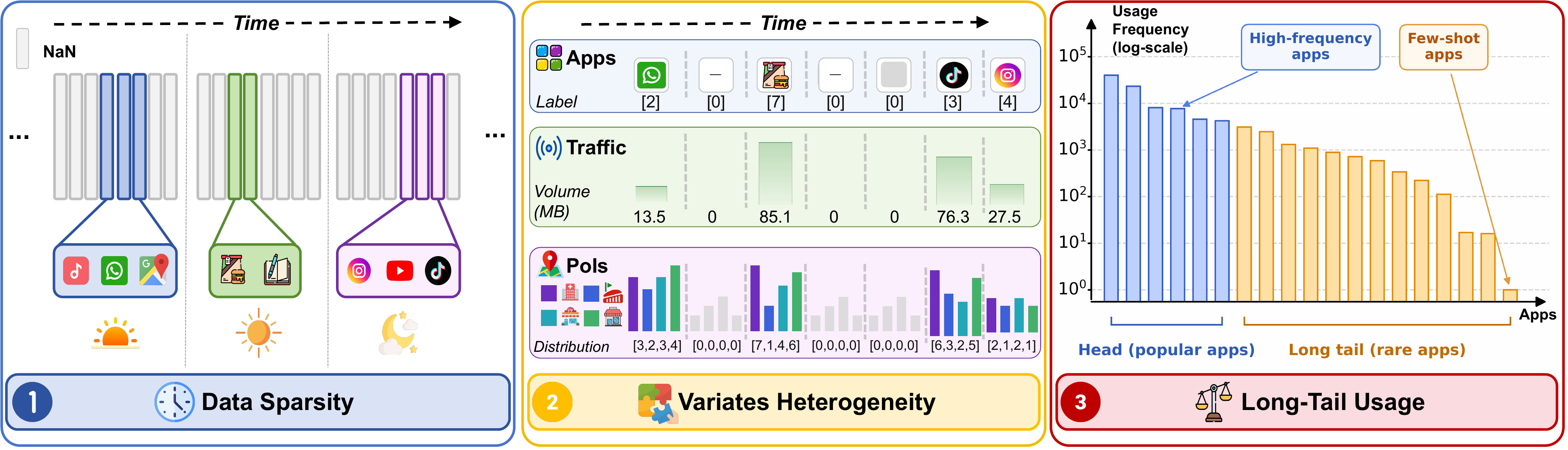}
  \caption{Three main challenges in mobile user trace generation.}
  \label{fig:challenges} % 整个图的引用标签
    \vspace{-0.5em}
\end{figure*}
Recent advances in mobile data generation mainly focus on injecting domain knowledge and information from extra modalities. 
KE-GAN~\cite{KEGAN} introduces a knowledge-enhanced generative framework that integrates semantic urban knowledge graphs to improve generation fidelity at the cellular level.
STK-Diff~\cite{STDIFF} adopts a diffusion-based approach by integrating spatial graphs, points of interest (POIs), and urban knowledge graphs into a structured denoising process, enabling controllable generation of cellular-level traffic volume with fine-grained spatial correlations. 
Nevertheless, both KE-GAN and STK-Diff operate at the cellular level, and thus cannot directly capture the sparsity and personalization present in user-level traces. 
NetDiff~\cite{netDiff} adopts a service-guided hierarchical diffusion model to generate app and traffic usage traces, but does not incorporate spatial context. 
AppGen~\cite{APPGEN} conditions app-sequence generation on mobility trajectories, but it focuses on app transitions and does not model the concurrent traffic volume.

Overall, existing methods fail to jointly address temporal sparsity and cross-channel heterogeneity in user-level mobile traces. Their interaction further exposes a long-tail usage imbalance: sparse activity makes frequently used apps and routine locations dominate the data, while heterogeneous variates make rare categories harder to model under a unified representation. These limitations highlight the need for a generative framework that enables fine-grained sparse reconstruction, decoupled heterogeneous modeling, and balanced representation of imbalanced categories.

To make these challenges explicit, we formalize mobile usage traces as three-dimensional data $\mathcal{Y} \in \mathbb{R}^{T \times M \times L}$, where $T$ indexes timesteps, $M$ indexes variate channels, and $L$ denotes the variate-specific semantic dimension. Specifically, for application and location channels, $L$ corresponds to discrete categories, while for traffic volume, it represents usage values. Under this abstraction, the challenges of modeling mobile usage traces can be summarized into three distinct problems, each corresponding to a subspace of $\mathcal{Y}$.

\begin{itemize}
    \item \textit{(C1) Temporal Sparsity in the Subspace $T$.} In real-world mobile usage, users interact with their devices intermittently rather than continuously, with activity often concentrated in short sessions such as commutes or breaks. Under regular sampling intervals, such sporadic behavior leads to severe temporal sparsity, making it difficult to extract effective local patterns~\cite{shukla2021multi, tipirneni2022self, chowdhury2023primenet}. The key challenge is therefore to preserve fidelity in both active and inactive periods: a generative model must accurately reconstruct usage values during active sessions while maintaining realistic usage frequencies to reflect the natural intermittency of human behavior.
   \item \textit{(C2) Cross-channel Heterogeneity across the Subspace $M$.} Mobile usage traces comprise heterogeneous variates, such as discrete application usage, discrete location states, and continuous traffic volume, which must be generated jointly. Applying the homogeneous modeling to these variates can induce gradient dominance~\cite{dualbalance,mtlcomb,heterogeneity}, where optimization is biased toward certain variate types while underrepresenting others. This necessitates a modeling approach that can capture variate-specific characteristics while preserving their joint dependencies.
    \item \textit{(C3) Long-tail Imbalance in the Subspace $L$.} Beyond temporal sparsity (C1) and cross-channel heterogeneity (C2), mobile usage traces exhibit pronounced long-tail imbalance across app categories. Apps serve diverse functional purposes and thus generate highly uneven usage patterns: a few high-engagement apps dominate most screen time, while many others are accessed only rarely~\cite{tail1,tail2}. The coexistence of sparsity and heterogeneity further amplifies this imbalance, making few-shot apps difficult to model reliably. Therefore, a generative model must not only capture the value distribution of rare apps, but also preserve the substantial differences in usage frequency across app categories.
\end{itemize}

These challenges underscore the necessity of a method that addresses the distinct demands of mobile usage traces. To this end, an effective solution must possess three core capabilities. Specifically, these are granular sensitivity to sparse behavioral signals, decoupled modeling mechanisms for heterogeneous variates, and spatially isolated representation with global calibration for imbalanced app usage.

In light of these requirements, we introduce the Cross-Gramian Angular Sum Field (C-GASF). This novel imaging method transforms complex multivariate time series including traffic volume, app categories, and spatial information into a unified image representation via phase cross-correlation. 
The C-GASF method is designed to address these three challenges through three corresponding encoding tracks.
Firstly, it enhances sensitivity to sparse user behaviors \textbf{C1} by mapping sparse activity events into structured image patterns that are easier for convolutional backbones to recover.
Secondly, to handle \textbf{C2} channel heterogeneity, it encodes app labels, traffic volume, and location clusters into separate but co-registered image components, reducing direct competition among heterogeneous channels. 
Thirdly, to address class imbalance \textbf{C3}, it assigns app categories to distinct spatial positions, allowing local convolutional filters to learn app-specific patterns while a global image distribution preserves overall frequency.

Building on the semantically separated C-GASF image, we propose Multivariate-Imaging Diffusion (MIDiff).
MIDiff is a diffusion framework whose U-Net backbone~\cite{Unet} is specifically augmented with Triplet Attention~\cite{Triple}. 
Triplet Attention factorizes attention along temporal, variable, and semantic axes, matching the structure of C-GASF images.

\begin{figure*}[!t] % [!ht] 是一个可选参数，让LaTeX尽量把图片放在这里 (Here) 或顶部 (Top)
  \centering
\includegraphics[width=1\textwidth]{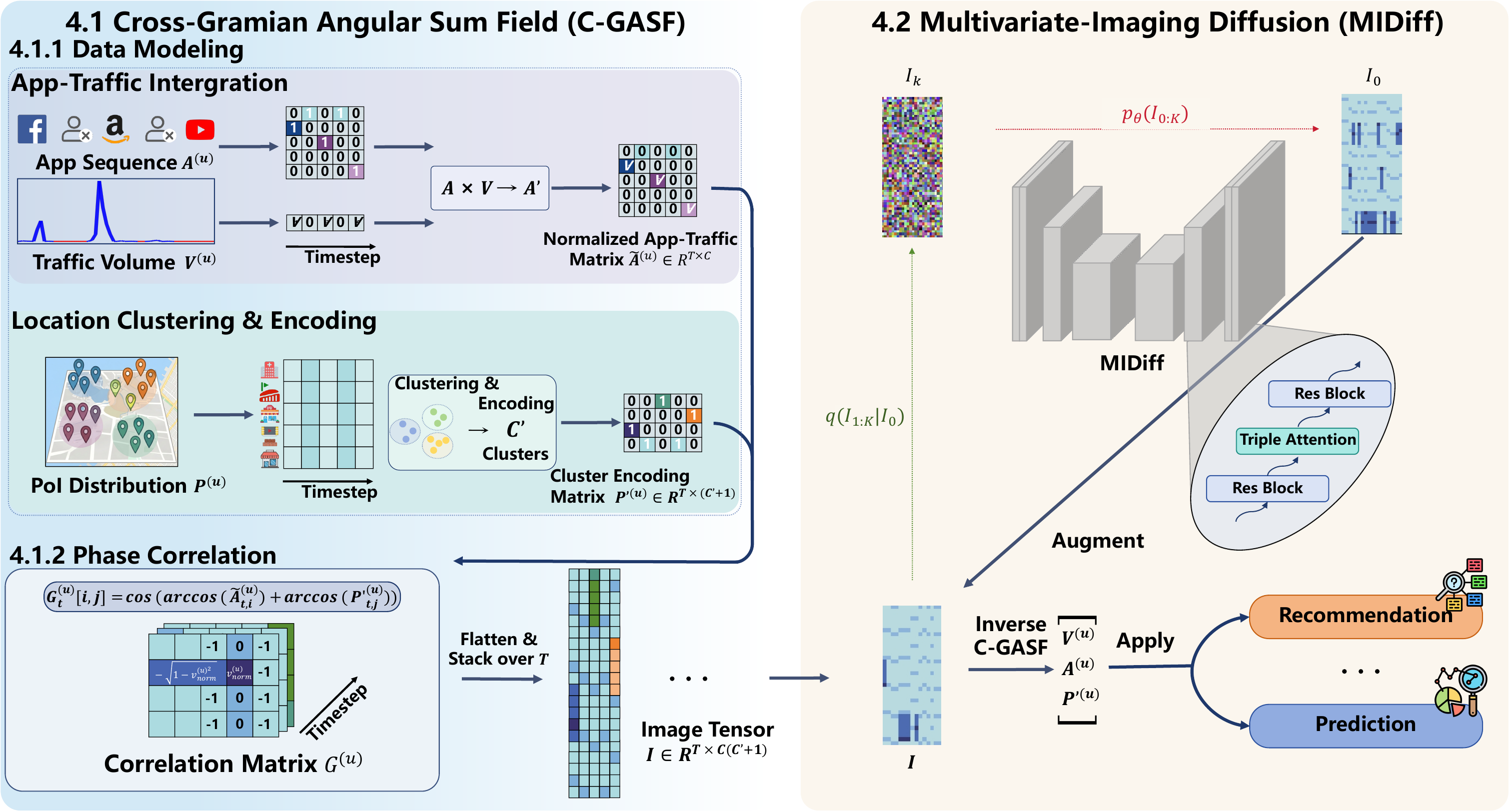}
  \caption{The framework of MIDiff.}
  \label{fig:overall_pipeline} % 整个图的引用标签
    \vspace{-0.5em}
\end{figure*}

This paper extends our earlier work\cite{poster}, by refining the problem formulation, representation, architecture, and experimental validation. The overall pipeline of our framework is illustrated in Figure~\ref{fig:overall_pipeline}, and the contributions of this work are summarized as follows,
\begin{itemize}
\item We propose C-GASF, which transforms sparse and heterogeneous mobile usage traces into a unified image representation. By encoding discrete app and spatial information as interest-point positions and continuous traffic volume as interest-point values, C-GASF decouples heterogeneous variates and enables CNNs to capture sparse app-specific usage patterns.

\item We introduce MIDiff, a diffusion generator built upon C-GASF for high-fidelity mobile usage trace synthesis. By augmenting the U-Net backbone with Triplet Attention, MIDiff effectively addresses data sparsity, channel heterogeneity, and app usage imbalance within a unified generative architecture.

\item We conduct comprehensive experiments on a real-world mobile usage dataset~\cite{dataset} against various baselines. Quantitative and qualitative results show that MIDiff consistently outperforms existing baselines and exhibits the three core capabilities required for mobile usage trace generation: fine-grained sparse reconstruction, decoupled heterogeneous modeling, and balanced representation of imbalance app usage.

\end{itemize}

\section{Related Work}

\subsection{General Time Series Generation}

\subsubsection{Sequence Modeling Methods.}
Sequence-based methods directly model temporal dependencies in 1D time series. TimeGAN~\cite{TimeGANStock} introduces a step-wise supervised loss to capture temporal dynamics, while TTS-GAN~\cite{TTSGAN} adopts a Transformer architecture to model long-range dependencies. COSCI-GAN~\cite{COSCI} employs channel-specific generators with a shared latent source to preserve inter-channel correlations. VAE-based approaches include VRAE~\cite{VRAE}, which combines VAEs with recurrent networks, TimeVAE~\cite{TimeVAE}, which explicitly models interpretable temporal components, and CR-VAE~\cite{CR-VAE}, which incorporates Granger causal structures.

More recently, diffusion models have achieved strong generative fidelity. Diffusion-TS~\cite{yuan2024diffusionts} uses an encoder-decoder Transformer to model temporal components, while PaD-TS~\cite{PaD} emphasizes the preservation of population-level properties. However, these general-purpose methods do not explicitly address the severe sparsity and variable heterogeneity of mobile usage traces.

\subsubsection{Image-Based Modeling Methods.}
Another line of work transforms 1D sequences into 2D representations and applies image-generation backbones. Gramian Angular Sum Field~\cite{GAF} converts a time series into an image by encoding temporal relationships in polar coordinates. Based on this representation, NetDiffus~\cite{netDiffus} applies diffusion models to GASF images of network traffic. ImagenTime~\cite{ImagenTime} instead combines the Short-Time Fourier Transform and delay embedding, while TimesNet~\cite{TimesNet} reshapes sequences into period-aware 2D tensors to capture intra- and inter-period variations.

Although these methods enable convolutional models to capture temporal patterns, their representations mainly emphasize dependencies within individual sequences and lack an explicit mechanism for modeling cross-variate correlations. MIDiff addresses this limitation through C-GASF, which jointly encodes multiple variables into a unified 2D correlation image.

\subsection{Mobile Traffic Generation}

\subsubsection{Cellular-Level Traffic Generation.}
Cellular-level methods generate traffic aggregated over base stations or geographic regions. KE-GAN~\cite{KEGAN} and ADAPTIVE~\cite{ADAPTIVE} incorporate urban knowledge graphs, with ADAPTIVE additionally using transfer learning for data-scarce regions. Recent diffusion approaches include STK-Diff~\cite{STDIFF}, which uses spatial graphs; OpenDiff~\cite{OpenDiff}, which replaces proprietary knowledge with public satellite and POI data; and STOUTER~\cite{STOUTER}, which constructs spatial and temporal graphs to model traffic fluctuations. However, aggregated cellular-level models cannot characterize individual user behavior.

\subsubsection{User-Level Traffic Generation.}
User-level methods are more closely related to our setting. MSH-GAN~\cite{MSHGAN} employs a multi-scale hierarchical GAN to model both individual and aggregate traffic patterns. NetDiff~\cite{netDiff} jointly generates app and traffic traces through a service-guided hierarchy, whereas PacketDiff~\cite{zhang2025packetdiff} models packet- and flow-level dependencies using graph-guided diffusion. LSDM~\cite{zhang2025lsdm} conditions app usage prediction on environmental text and satellite imagery, while AppGen~\cite{APPGEN} generates personalized app sequences conditioned on mobility trajectories.

Nevertheless, these methods model only subsets of user behavior: NetDiff and PacketDiff omit spatial context, LSDM does not model mobility, and AppGen does not jointly generate traffic volume. Moreover, existing approaches often avoid severe sparsity rather than modeling it directly. For example, MSH-GAN removes highly sparse records during preprocessing, potentially distorting the real usage-frequency distribution. In contrast, MIDiff retains sparse traces and uses C-GASF to amplify sporadic activities in the imaging space, facilitating the modeling of sparse and heterogeneous user behavior.

\vspace{-0.5em}
% \end{figure*}
\section{Preliminaries}
\subsection{Mobile Usage Traces}
Let $\mathcal{U}$ denote the set of users in the dataset. For each user $u\in\mathcal{U}$, the mobile usage trace contains user behaviors collected at fixed sampling intervals over $T$ timesteps, denoted as $X^{(u)} = \{x_t^{(u)}\}_{t=1}^T$. At each timestep $t$, the user's behavior $x_t^{(u)}$ is represented as a composite vector $\{v_t^{(u)}, a_t^{(u)}, p_t^{(u)}\}$. Here, $v_t^{(u)}$ is a continuous variable representing the network traffic volume consumed by the user. $a_t^{(u)}$ represents app usage, encoded as a discrete label indicating the app category in use.  $p_t^{(u)}$ denotes the POI count vector associated with the user's location, represented as
$p_t^{(u)}=[p_{t,1}^{(u)},p_{t,2}^{(u)},\dots,p_{t,O}^{(u)}]$.
Here, each element $p_{t,o}^{(u)}\in\mathbb{N}$ represents the count of a specific POI category, such as shops and parks, around the base station, and $O$ is the total number of POI categories.

\subsection{Gramian Angular Sum Field}
Given a univariate time series $S=\{s_1,\dots,s_T\}$, GASF first applies min-max normalization to rescale each value into $[0,1]$,
\begin{equation}
\tilde{s}_i = \frac{s_i-\min(S)}{\max(S)-\min(S)},
\end{equation}
where $\tilde{s}_i\in[0,1]$. The normalized value is then represented in the
polar coordinate system by encoding its magnitude as an angle
\begin{equation}
\phi_i = \arccos(\tilde{s}_i).
\end{equation}
GASF is defined by computing the trigonometric sum
between every pair of angular values
\begin{equation}
\mathrm{GASF}_{i,j}
=
\cos(\phi_i+\phi_j).
\end{equation}

In the matrix, the main diagonal preserves the self-angular information of each timestamp:
\begin{equation}
\mathrm{GASF}_{i,i}
=
\cos(2\phi_i)
=
2\tilde{s}_i^2-1.
\end{equation}
Therefore, under $\tilde{x}_i\in[0,1]$, the
normalized value can be recovered from the diagonal by
\begin{equation}
\tilde{s}_i
=
\sqrt{\frac{\mathrm{GASF}_{i,i}+1}{2}}.
\end{equation}
The original value can then be obtained by applying the inverse normalization:
\begin{equation}
s_i
=
\tilde{s}_i\left(\max(S)-\min(S)\right)+\min(S).
\end{equation}

\section{METHODOLOGY}
\begin{figure*}[!t] % [!ht] 是一个可选参数，让LaTeX尽量把图片放在这里 (Here) 或顶部 (Top)
  \centering
\includegraphics[width=1\textwidth, ]{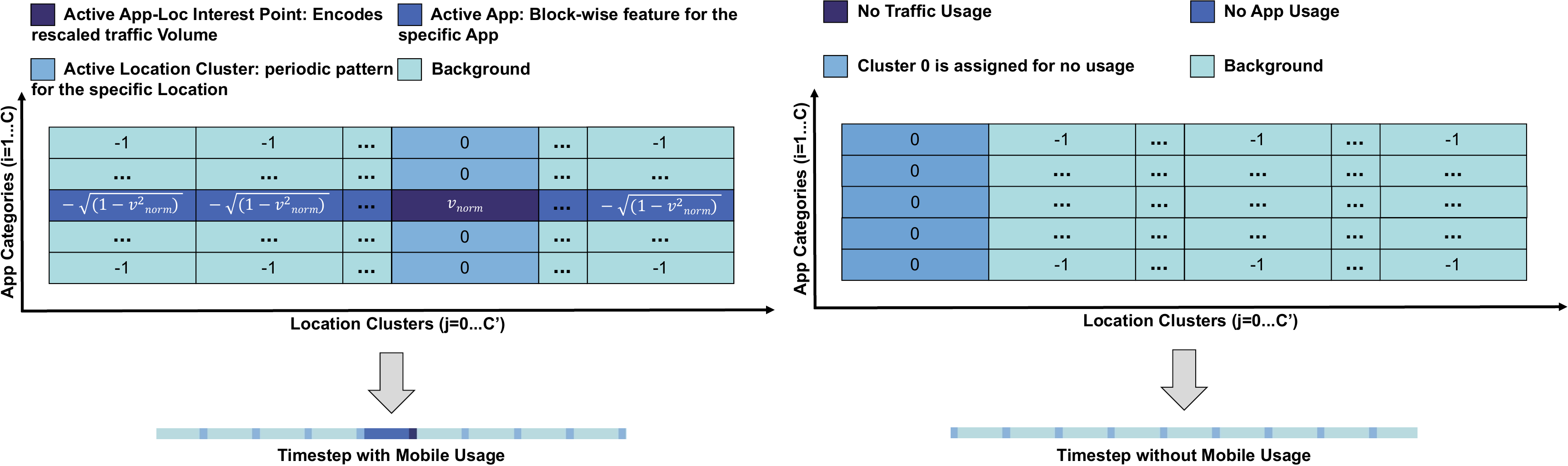}

  \caption{The illustration of C-GASF image. Each row contains the information of the user's mobile usage trace in one timestep.}
  \label{fig:C-GASF} % 整个图的引用标签
    \vspace{-0.5em}
\end{figure*}
\subsection{Cross-Gramian Angular Sum Field}
\subsubsection{Data Modeling}
To address the challenges of sparsity and complex interdependencies in the mobile usage traces, we transform the raw traffic volumes, app usage sequences, and location data as
$V^{(u)}=\{v_1^{(u)},v_2^{(u)},\dots,v_T^{(u)}\}$,
$A^{(u)}=\{a_1^{(u)},a_2^{(u)},\dots,a_T^{(u)}\}$, and
$P^{(u)}=\{p_1^{(u)},p_2^{(u)},\dots,p_T^{(u)}\}$, respectively.
The superscript $(u)$ indicates that the corresponding trace belongs to user $u$.
We transform these raw sequences into structured representations with explicit consistency constraints.

\textbf{App-Traffic Integration.}
First, the app usage sequence $A^{(u)} = \{a_1^{(u)}, a_2^{(u)}, \dots, a_T^{(u)}\}$, where each $a_t^{(u)} \in \{1, \dots, C\}$ is the app category, is converted into a sequence of one-hot encoded vectors $H^{(u)} = \{h_1^{(u)}, h_2^{(u)}, \dots, h_T^{(u)}\}$, where each $h_t^{(u)} \in \{0, 1\}^C$. Subsequently, this one-hot encoded app usage $H^{(u)}$ and the traffic volumes $V^{(u)} = \{v_1^{(u)}, v_2^{(u)}, \dots, v_T^{(u)}\}$ are merged into a unified matrix $\mathbf{A'^{(u)}} \in \mathbb{R}^{T \times C}$. For each timestep $t \in \{1, \dots, T\}$, the $t$-th row of $\mathbf{A'^{(u)}}$ is constructed by multiplying the one-hot vector $h^{(u)}_t$ with the corresponding traffic volume $v^{(u)}_t$,
\begin{equation}
A'^{(u)}_t = h^{(u)}_t \times v^{(u)}_t.\end{equation}
This means, for each element $A'^{(u)}_{t,i}$ where $t \in \{1,\dots,T\}$ and category $i \in \{1,\dots,C\}$,

\begin{equation}
\mathbf{A'^{(u)}}_{t,i} = 
\begin{cases} 
v^{(u)}_t, & \text{if app }i\text{ is active at }t\text{ with traffic }v^{(u)}_t, \\
0, & \text{otherwise}.
\end{cases}
\end{equation}

Similar to the preprocessing required for GASF, we first rescale the traffic volumes into $[0,1]$. For each app category $i$, we normalize its corresponding column in $\mathbf{A'^{(u)}}$ by the maximum observed traffic volume for that specific category across the entire dataset,
    \begin{equation}
    \tilde{\mathbf{A}}_{t,i}^{(u)}
    =
    \frac{\mathbf{A'}_{t,i}^{(u)}}{
    \displaystyle\max_{\upsilon \in U,\ \tau \in \{1,\dots,T\}}
    \mathbf{A'}_{\tau,i}^{(\upsilon)}
    },
\end{equation}
where the denominator denotes the global maximum traffic volume for app category $i$ across all users and timesteps in the dataset. This normalization maps the traffic volumes $\mathbf{A'^{(u)}}_{t,i}$ from $[0,+\infty)$ to $[0,1]$.

\begin{figure*}[!t] % [!ht] 是一个可选参数，让LaTeX尽量把图片放在这里 (Here) 或顶部 (Top)
  \centering
\includegraphics[width=\textwidth,height=0.15\textheight]{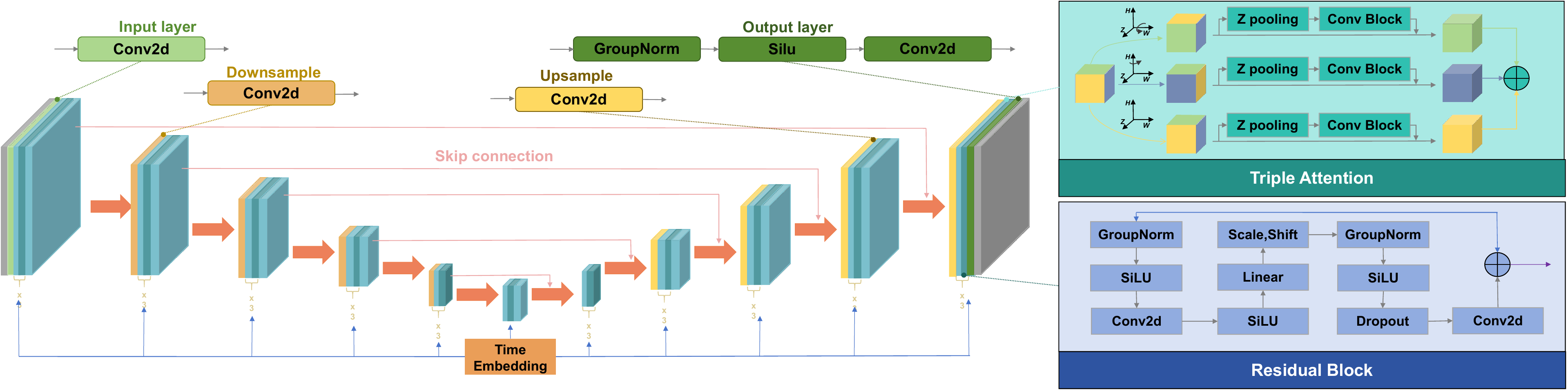}
  \caption{MIDiff model architecture.}
  \label{fig:Unet} % 整个图的引用标签
    \vspace{-0.5em}
\end{figure*}
\textbf{Location Clustering and Encoding.}
\label{clustering}
The location component $P$ consists of high-dimensional location vectors representing counts of different POI types, such as transportation hubs, commercial centers, at each location. To enable effective modeling and joint representation with app-traffic data, we first apply K-means clustering to these location vectors, grouping them into $C'$ distinct spatial-context patterns. And subsequently we encode cluster assignments into a one-hot matrix $\mathbf{P'^{(u)}} \in \{0,1\}^{T\times(C'+1)}$,
\begin{equation}
\mathbf{P}^{\prime(u)}_{t,j}
=
\begin{cases}
1, &
\resizebox{0.3\textwidth}{!}{$\left(j=0 \land \tilde{\mathbf{A}}^{(u)}_{t,:}=\mathbf{0}\right) \lor
\left(j=g(p_t^{(u)}) \land \tilde{\mathbf{A}}^{(u)}_{t,:}\neq\mathbf{0}\right),$}\\
0, & \text{otherwise},
\end{cases}
\end{equation}

where $t \in \{1,\dots,T\}$, $j\in\{0,1,\dots,C'\}$ and $g(p_t^{(u)})\in\{1,\dots,C'\}$ denotes the cluster label for $p_t^{(u)}$.

To explicitly model the data sparsity of mobile data, we assign timesteps without user behavior to the $0$-index cluster. This dedicated representation enhances semantic clarity and facilitates the inverse transformation process.
To formalize this distinction between active and inactive steps, we define a behavior indicator vector $\boldsymbol{\delta} \in \{0,1\}^T$,
\begin{equation}
\delta_t = \mathbb{I}\!\left(\sum_{i=1}^C \tilde{\mathbf{A}}_{t,i}^{(u)} > 0\right).
\end{equation}

\begin{itemize}
\item In the inactive case where $\delta_t = 0$, the $t$-th row of $\tilde{\mathbf{A}}^{(u)}$ is set to a zero vector. Correspondingly, the $0$-th entry of $\mathbf{P'}^{(u)}$ is set to one, such that $\mathbf{P'}_{t,0}^{(u)} = 1$.
\item In the active case where $\delta_t = 1$, the $t$-th row of $\tilde{\mathbf{A}}^{(u)}$ must contain exactly one non-zero entry representing the active app category and its traffic volume. Simultaneously, the $t$-th row of $\mathbf{P'}^{(u)}$ must specify the corresponding location cluster by having exactly one entry equal to $1$ at an index $j \geq 1$.
\end{itemize}
These guarantees ensure a coherent spatial-temporal app-traffic relationships within the transformed representation $(\tilde{\mathbf{A}}^{(u)}, \mathbf{P'}^{(u)}, \boldsymbol{\delta})$. The constraint of having at most one non-zero entry in $\tilde{\mathbf{A}}^{(u)}$ at each active timestep is crucial, as it makes the inverse transformation possible for the subsequent steps.

After preprocessing, each sample is represented by the pair $(\tilde{\mathbf{A}}^{(u)},\mathbf{P'}^{(u)})$. We now transform these matrices into a unified tensor that preserves both spatial–temporal structure and cross-modal correlations.
\vspace{-0.5em}
\subsubsection{Phase Correlation}
\label{C-GASF}

To couple app usage and location context, we compute in a way similar to GASF. At each step $t$, the normalized vectors $\tilde{\mathbf{A}}_{t}^{(u)}$ and $\mathbf{P'}_{t,\,:}^{(u)}$ are interpreted as cosine coordinates and combined by,  
\begin{equation}
\mathbf{G}_t^{(u)}[i,j]=\cos\!\bigl(\arccos(\tilde{\mathbf{A}}_{t,i}^{(u)})+\arccos(\mathbf{P'}_{t,j}^{(u)})\bigr),
\end{equation}  
which simplifies to,
\begin{equation}
\mathbf{G}_t^{(u)}[i,j]=\tilde{\mathbf{A}}_{t,i}^{(u)}\mathbf{P'}_{t,j}^{(u)}-\sqrt{1-\tilde{\mathbf{A}}_{t,i}^{(u)^{2}}}\sqrt{1-\mathbf{P'}_{t,j}^{(u)^{2}}}.
\end{equation}
Following the consistency conditions already enforced on $(\mathbf{A}^{(u)},\mathbf{P}^{(u)})$, two cases arise,
      
\begin{itemize}
\item \textbf{No behaviour} ($\delta_t=0$), $\mathbf{P'}^{(u)}_{t,0}=1$ and $\mathbf{P'}^{(u)}_{t,j}=0$ for $j\ge 1$, giving,
\begin{equation}
\mathbf{G}_t^{(u)}[i,j]=
\begin{cases}
0, & j=0,\\
-1, & j\ge 1.
\end{cases}
\end{equation}

\item \textbf{Behaviour present} ($\delta_t=1$), letting $i_t$ and $j_t$ denote the unique active app and POIs cluster in each timestep, and $v_{\mathrm{norm}}^{(u)}=\tilde{\mathbf{A}}_{t,i_t}^{(u)}$, we derive a pattern that encodes app information into block structures and location information into period columns, where the traffic volume is determined by their intersection,
\begin{equation}
\mathbf{G}_t^{(u)}[i,j]=
\begin{cases}
v_{\mathrm{norm}}^{(u)}, & i=i_t,\,j=j_t,\\
-\sqrt{1-v_{\mathrm{norm}}^{(u)^{2}}}, & i=i_t,\,j\neq j_t,\\
0, & i\neq i_t,\,j=j_t,\\
-1, & \text{otherwise}.
\end{cases}
\end{equation}
\end{itemize}
Here, $\mathbf{G}_t^{(u)}[i,j]$ denotes the element in the $i$-th row (corresponding to the $i$-th app category) and $j$-th column (corresponding to the $j$-th location cluster including the $0$-index for inactivity) in the correlation matrix at time $t$. The resulting matrix $\mathbf{G}^{(u)}_t \in \mathbb{R}^{C \times (C'+1)}$ encodes the interaction between app usage and spatial context at each timestep.

Finally, each matrix $\mathbf{G}_t^{(u)}$ is row-major vectorized into $\mathbf{g}_t^{(u)}=\mathrm{vec}(\mathbf{G}_t^{(u)})\in\mathbb{R}^{C(C'+1)}$.  Concatenating all timesteps yields the compact representation,  
\begin{equation}
\mathbf{I^{(u)}}= \begin{bmatrix}{\mathbf{g}_1^{(u)}}^\top\\ \vdots\\ {\mathbf{g}_T^{(u)}}^\top\end{bmatrix}\in\mathbb{R}^{T\times C(C'+1)},
\end{equation}  
Thus, the overall transform yields an image in which three heterogeneous variates—app category, traffic volume, and POIs cluster—are jointly represented.  
Each positive entry $\mathbf{I}_{t,m}^{(u)}$ encodes the rescaled traffic volume at the corresponding ``interest point'' $(i_t,j_t)$, while the remaining positions exhibit a regular pattern,  
all entries with $i=i_t$ but $j\neq j_t$ carry $-\sqrt{1-v_{\mathrm{norm}}^{2}}$, those with $j=j_t$ but $i\neq i_t$ are set to $0$, and the rest are fixed at $-1$.  
This deliberate geometric layout exposes salient patterns that convolutional architectures can capture effortlessly. Substantially, it alleviates the difficulty of representation learning, as illustrated in Figure \ref{fig:C-GASF}. 

From a computational perspective, for a single usage trace of length $T$ with $C$ app categories and $P$ POI clusters, the transformation requires $\mathcal{O}(TCP)$ time and $\mathcal{O}(TCP)$ space, scaling linearly with respect to each structural dimension. Meanwhile, its block-wise app features are also significant for recovering the original mobile usage trace from the C-GASF image. The complete inverse transformation algorithm is detailed in the supplementary material.

\subsection{Multivariate-Imaging Diffusion}

The image $\mathbf{I}\in\mathbb{R}^{T\times C(C'+1)}$, equivalently $\mathbf{I}\in\mathbb{R}^{1\times H\times W}$ with $H=T$ and $W=C(C'+1)$, produced in Sec.~\ref{C-GASF} is interpreted as an image tensor. It possesses a unique structure, where its W-axis encodes behavioral information and block structures at a single timestep, while the H-axis  represents aggregate information for a single app or location across all timesteps. We learn its distribution with MIDiff $\boldsymbol{\epsilon}_{\theta}$, an unconditional diffusion model with U-Net backbone.

\textbf{U-Net architecture.}
The network follows the standard encoder–decoder design with residual blocks. For the input feature at current layer with dimension $z \times h \times w$. To effectively learn the underlying imaging mechanism and behavioral distributions, a mechanism capable of decoupling information along the H and W-axes is required. We therefore replace the standard attention blocks with \emph{Triplet Attention}~\cite{Triple}. This mechanism factorizes the computation into three branches. The \emph{H-axis aggregation} permutes the tensor to $\mathbb{R}^{h \times z \times w}$ and applies a Spatial-Gate pooling along the $H$ (temporal) axis, which is responsible for decoupling modeling of individual apps and locations. Concurrently, the \emph{W-axis aggregation} permutes $I$ to $\mathbb{R}^{w \times h \times z}$ and pools along the $W$ (app/location) axis, ensuring the modeling of patterns within a single timestep. Finally, the \emph{$Z$-axis aggregation} (the spatial branch) applies a Spatial-Gate with ChannelPool along the $Z$ (channel) dimension, allowing the model to learn the user's overall behavior. The outputs of the H and W branches are permuted back to $\mathbb{R}^{z\times h\times w}$, and the results of all three branches are combined and residually added to the input $I$. This decoupled, multi-axis approach allows the network to effectively learn the distinct distributions along the temporal, modality, and channel information. The overall architecture of the U-Net is shown in Figure \ref{fig:Unet}.

\textbf{Forward Noising Process.}
Given a clean C-GASF image $\mathbf{I}_0$ sampled from the data distribution, 
we define a Markov forward noising process $q$ that progressively adds Gaussian 
noise over $K$ discrete steps
\begin{equation}
q(\mathbf{I}_{1:K}|\mathbf{I}_0)
=
\prod_{k=1}^{K} q(\mathbf{I}_k|\mathbf{I}_{k-1}).
\end{equation}
At each step, the transition is defined as
\begin{equation}
q(\mathbf{I}_k|\mathbf{I}_{k-1})
=
\mathcal{N}\!\left(
\sqrt{\alpha_k}\mathbf{I}_{k-1},
\beta_k\mathbf{I}
\right),
\end{equation}
where $\alpha_k=1-\beta_k$. Equivalently, the noised image at an arbitrary 
step $k$ can be sampled directly from $\mathbf{I}_0$ as
\begin{equation}
\mathbf{I}_k
=
\sqrt{\bar{\alpha}_k}\,\mathbf{I}_0
+
\sqrt{1-\bar{\alpha}_k}\,\boldsymbol{\epsilon},
\quad
\boldsymbol{\epsilon} \sim \mathcal{N}(\mathbf{0}, \mathbf{I}),
\end{equation}
where $\bar{\alpha}_k = \prod_{i=1}^{k}\alpha_i$, following the cosine 
schedule~\cite{Nichol2021ImprovedDD}.

\textbf{Reverse Denoising Process.}
The generative capability lies in learning to reverse the corruption. Starting 
from pure noise $\mathbf{I}_K \sim \mathcal{N}(\mathbf{0}, \mathbf{I})$, we 
define a learnable reverse denoising process $p_\theta$:
\begin{equation}
p_\theta(\mathbf{I}_{0:K})
=
p(\mathbf{I}_K)
\prod_{k=1}^{K} p_\theta(\mathbf{I}_{k-1}|\mathbf{I}_k).
\end{equation}
At each reverse step, MIDiff predicts the noise component 
$\boldsymbol{\epsilon}$ from the noised image $\mathbf{I}_k$ and timestep $k$, 
and the transition can be written as
\begin{equation}
p_\theta(\mathbf{I}_{k-1}|\mathbf{I}_k)
=
\mathcal{N}\!\left(
\mathbf{I}_{k-1};
\boldsymbol{\mu}_\theta(\mathbf{I}_k,k),
\sigma_k^2\mathbf{I}
\right),
\end{equation}
where
\begin{equation}
\boldsymbol{\mu}_\theta(\mathbf{I}_k,k)
=
\frac{1}{\sqrt{\alpha_k}}
\left(
\mathbf{I}_k
-
\frac{1-\alpha_k}{\sqrt{1-\bar{\alpha}_k}}
\boldsymbol{\epsilon}_\theta(\mathbf{I}_k,k)
\right).
\end{equation}
Thus, each denoising step can be sampled as
\begin{equation}
\mathbf{I}_{k-1}
=
\boldsymbol{\mu}_\theta(\mathbf{I}_k,k)
+
\sigma_k\mathbf{z},
\quad
\mathbf{z} \sim \mathcal{N}(\mathbf{0}, \mathbf{I}) \ \text{if } k>1.
\end{equation}

\textbf{Training Objective.}
The network is optimized by minimizing the simple Mean Square Error (MSE) loss,
\begin{equation}
\mathcal{L}_{\text{simple}}(\theta)=\mathbb{E}_{\mathbf{I}_0,\boldsymbol{\epsilon},k}\!\left[\lVert\boldsymbol{\epsilon}-\boldsymbol{\epsilon}_{\theta}(\mathbf{I}_{k},k)\rVert^{2}\right],
\end{equation}
where $\mathbf{I}_0$ is sampled from the training set. By accurately predicting the noise component that corrupts the structured behavioral information in $\mathbf{I}_0$, the model learns to reverse the noising process and generate C-GASF images with real distributions of interest points in the image domain. Ultimately, this capability translates into the ability to model sparse mobile usage, capturing patterns of rarely used apps and the multi-variate correlations in mobile usage traces.
\vspace{-1em}

\section{Experiments}

\subsection{Experiment Setting}
\textbf{Implementations.}
We conduct our experiments using \textbf{App Usage Dataset}~\cite{dataset}, which was collected over one week in one of the largest cities in China. The raw records are first aggregated into 15-minute intervals. We then construct fixed-length samples by segmenting each user's weekly trace into two-day windows. Since each day contains 96 intervals under the 15-minute granularity, each sample has a sequence length of $T=192$. The details of the dataset are summarized in Table~\ref{tab:dataset_stats}. As described in Section~\ref{clustering}, we apply k-means to the high-dimensional location data to reduce the POIs of different base stations into six discrete cluster labels, plus an additional label to denote inactivity timesteps. We report computational cost in Table~\ref{tab:diffusion_efficiency_bs2}.

\begin{table}[t]
\centering
\caption{Statistics of \textbf{App Usage Dataset}.} % 建议修改标题以匹配内容
\label{tab:dataset_stats}
\small 
\begin{tabular*}{0.8\linewidth}{@{\extracolsep{\fill}}cc} 
\toprule
Dataset Statistic & Value \\
\midrule
Duration & One week \\
% 修正语法
Number of identified Apps & 2000 \\ % 增加千分位符
Number of users & 1000 \\
Number of App categories & 20 \\ % 修正语法
Number of POI categories & 17 \\
\bottomrule
\end{tabular*}
  \vspace{-0.5em}
\end{table}

\begin{table}[!ht]
\centering
\caption{Sampling efficiency comparison per denoising step among diffusion-based models under FP32 precision and batch size equal to 2 on single A100.}
\label{tab:diffusion_efficiency_bs2}
\small
\setlength{\tabcolsep}{5pt}
\resizebox{\linewidth}{!}{
\begin{tabular}{lccccc}
\toprule
Model 
& Params (M) 
& GFLOPs 
& Latency (ms) 
& FPS 
& Peak Mem. (MB) \\
\midrule
Diffusion-TS
& 0.52
& 0.19
& 9.24
& 216.46
& 13.92 \\

PaD-TS
& 9.14
& 1.24
& 6.11
& 327.47
& 47.89 \\

ImagenTime
& 142.05
& 1790.44
& 133.10
& 15.03
& 2197.38 \\

TimeAutoDiff
& 24.19
& 0.72
& 14.45
& 138.37
& 193.69 \\
\midrule
MIDiff
& 148.53
& 413.14
& 66.68
& 30.00
& 940.81 \\
\bottomrule
\end{tabular}}
\vspace{0.5mm}
\end{table}
\textbf{Baselines.}
We select 9 general time series generative models, ranging from seminal classic works to recent challenge-oriented approaches. 
(1) Classic approaches include TTS-GAN~\cite{TTSGAN} and TimeGAN~\cite{TimeGANStock}, which extend adversarial learning to time series generation by incorporating temporal inductive biases into the GAN framework.
(2) Diffusion-based approaches for multivariate generation include Diffusion-TS~\cite{yuan2024diffusionts} and PaD-TS~\cite{PaD}, which improve the fidelity of generated multivariate time series by incorporating temporal decomposition, frequency-domain constraints, or population-level dependency preservation.
(3) Image-based approaches include ImagenTime~\cite{ImagenTime}, which transforms time series into image representations via delay embedding or STFT and leverages vision diffusion models for generation.
(4) Approaches for variate heterogeneity include TimeAutoDiff~\cite{TimeAutoDiff}, which uses a VAE to encode mixed-type time-series tabular features into a continuous latent space for latent diffusion modeling.
(5) Approaches for data sparsity include ZITS~\cite{zits}, which decouples zero-inflated generation into Bernoulli-gated occurrence modeling and non-zero magnitude estimation.
\begin{table*}[!t]
  \centering
  \caption{Generation results on \textbf{App Usage Dataset}. Green and gray values in the Gain columns denote MIDiff's improvement or degradation relative to each baseline. For distance metrics, Gain is reported as relative improvement with respect to the baseline score, while for DA it is reported as absolute difference.}
  \label{tab:scorew}
  \begin{threeparttable}
    % Requires:
    %\usepackage{booktabs}
    % \usepackage{multirow}
    %\usepackage[table]{xcolor}
    %\usepackage{nicematrix}

    \definecolor{posGreen}{RGB}{50,130,90}
    \definecolor{topGray}{RGB}{242,242,242}
    \definecolor{midiffBlue}{RGB}{235,245,255}

    \newcommand\B[1]{{\bfseries #1}}
    \newcommand\U[1]{\underline{#1}}
    \newcommand\ImpP[1]{\textcolor{posGreen}{\scriptsize #1}}
    \newcommand\ImpN[1]{\textcolor{gray}{\scriptsize #1}}

    {\small
    \setlength{\tabcolsep}{0pt}
    \renewcommand{\arraystretch}{1.08}
    \newlength{\scoredeltasep}
    \setlength{\scoredeltasep}{1pt}

    \begin{NiceTabular*}{\linewidth}{
    @{\extracolsep{\fill}} l
    @{\extracolsep{\fill}} l
    @{\extracolsep{\fill}} c@{\extracolsep{1pt}\hspace{\scoredeltasep}}c
    @{\extracolsep{\fill}} c@{\extracolsep{1pt}\hspace{\scoredeltasep}}c
    @{\extracolsep{\fill}} c@{\extracolsep{1pt}\hspace{\scoredeltasep}}c
    @{\extracolsep{\fill}} c@{\extracolsep{1pt}\hspace{\scoredeltasep}}c
    }[cell-space-limits=0pt]
      \CodeBefore
        \rectanglecolor{topGray}{1-1}{2-10}
        \rectanglecolor{midiffBlue}{11-1}{11-10}
      \Body
      \toprule
      \multirow{2}{*}{Type} & \multirow{2}{*}{Baseline} 
      & \multicolumn{2}{c}{VDS $\downarrow$} 
      & \multicolumn{2}{c}{FDDS $\downarrow$} 
      & \multicolumn{2}{c}{DA $\downarrow[0,0.5]$} 
      & \multicolumn{2}{c}{Predictive Score $\downarrow$} \\
      \cmidrule(lr){3-4} \cmidrule(lr){5-6} \cmidrule(lr){7-8} \cmidrule(lr){9-10}
      & & Score & Gain 
        & Score & Gain 
        & Score & Gain 
        & Score & Gain \\
      \midrule

      \multirow{2}{*}{\textit{Classic}}
      & TimeGAN      & 0.0553 & \ImpP{+95.30\%} & 0.6092 & \ImpP{+95.19\%} & 0.4953 & \ImpP{+0.3427} & 19404   & \ImpP{+62.40\%} \\
      & TTS-GAN      & 0.1104 & \ImpP{+97.64\%} & 0.1280 & \ImpP{+77.11\%} & 0.4976 & \ImpP{+0.3450} & 43016   & \ImpP{+83.04\%} \\

      \addlinespace[0.5pt]

      \multirow{2}{*}{\textit{Variate Correlation}}
      & Diffusion-TS & 0.0856 & \ImpP{+96.96\%} & \U{0.0352} & \ImpP{+16.76\%} & 0.3511 & \ImpP{+0.1985} & 7331    & \ImpP{+0.48\%} \\
      & PaD-TS       & 0.5636 & \ImpP{+99.54\%} & 0.2585 & \ImpP{+88.67\%} & 0.5000 & \ImpP{+0.3474} & 1123510 & \ImpP{+99.35\%} \\

      \addlinespace[0.5pt]

      \multirow{1}{*}{\textit{Imaging}}
      & ImagenTime   & 0.0715 & \ImpP{+96.36\%} & 0.0785 & \ImpP{+62.68\%} & 0.4907 & \ImpP{+0.3381} & 47536   & \ImpP{+84.65\%} \\

      \addlinespace[0.5pt]

      \textit{Heterogeneous}
      & TimeAutoDiff & 0.1016 & \ImpP{+97.44\%} & 0.3166 & \ImpP{+90.75\%} & 0.3892 & \ImpP{+0.2366} & 7138    & \ImpN{-2.21\%} \\

      \addlinespace[0.5pt]

      \multirow{2}{*}{\textit{Sparse}}
      & ZITS-GAN     & 0.0223 & \ImpP{+88.34\%} & 0.2052 & \ImpP{+85.72\%} & 0.4646 & \ImpP{+0.3120} & \U{6829} & \ImpN{-6.84\%} \\
      & ZITS-VAE     & \U{0.0214} & \ImpP{+87.85\%} & 0.1036 & \ImpP{+71.72\%} & \U{0.3476} & \ImpP{+0.1950} & \B{6731} & \ImpN{-8.39\%} \\

      \addlinespace[0.5pt]
      \midrule

      \textit{Ours}
      & MIDiff       & \B{0.0026} & --- & \B{0.0293} & --- & \B{0.1526} & --- & 7296 & --- \\
      \bottomrule
    \end{NiceTabular*}
    }
  \end{threeparttable}
\end{table*}

\begin{table*}[!t]
  \centering
  \caption{Feature and distance-based measures on TS-Bench between generated datasets and \textbf{App Usage Dataset}.}
  \label{tab:results_table3_all}
  \begin{threeparttable}
    % Requires:
    %\usepackage{booktabs}
    % \usepackage{multirow}
    %\usepackage[table]{xcolor}
    %\usepackage{nicematrix}

    \definecolor{posGreen}{RGB}{50,130,90}
    \definecolor{topGray}{RGB}{242,242,242}
    \definecolor{midiffBlue}{RGB}{235,245,255}

    \newcommand\B[1]{{\bfseries #1}}
    \newcommand\U[1]{\underline{#1}}
    \newcommand\ImpP[1]{\textcolor{posGreen}{\scriptsize #1}}
    \newcommand\ImpN[1]{\textcolor{gray}{\scriptsize #1}}

    {\scriptsize
    \setlength{\tabcolsep}{0pt}
    \renewcommand{\arraystretch}{1.08}
    \newlength{\scoredeltasepFeature}
    \setlength{\scoredeltasepFeature}{1pt}

    \begin{NiceTabular*}{\linewidth}{
    @{\extracolsep{\fill}} l
    @{\extracolsep{\fill}} c@{\extracolsep{1pt}\hspace{\scoredeltasepFeature}}c
    @{\extracolsep{\fill}} c@{\extracolsep{1pt}\hspace{\scoredeltasepFeature}}c
    @{\extracolsep{\fill}} c@{\extracolsep{1pt}\hspace{\scoredeltasepFeature}}c
    @{\extracolsep{\fill}} c@{\extracolsep{1pt}\hspace{\scoredeltasepFeature}}c
    @{\extracolsep{\fill}} c@{\extracolsep{1pt}\hspace{\scoredeltasepFeature}}c
    @{\extracolsep{\fill}} c@{\extracolsep{1pt}\hspace{\scoredeltasepFeature}}c
    }[cell-space-limits=0pt]
      \CodeBefore
        \rectanglecolor{topGray}{1-1}{2-13}
        \rectanglecolor{midiffBlue}{11-1}{11-13}
      \Body
      \toprule
      \multirow{2}{*}{Baseline}
      & \multicolumn{2}{c}{MDD $\downarrow$}
      & \multicolumn{2}{c}{ACD $\downarrow$}
      & \multicolumn{2}{c}{SD $\downarrow$}
      & \multicolumn{2}{c}{KD $\downarrow$}
      & \multicolumn{2}{c}{DTW $\downarrow$}
      & \multicolumn{2}{c}{ED $\downarrow$} \\
      \cmidrule(lr){2-3} \cmidrule(lr){4-5} \cmidrule(lr){6-7}
      \cmidrule(lr){8-9} \cmidrule(lr){10-11} \cmidrule(lr){12-13}
       & Score & Gain
       & Score & Gain
       & Score & Gain
       & Score & Gain
       & Score & Gain
       & Score & Gain \\
      \midrule

      TimeGAN      & 43500.76 & \ImpP{+88.27\%} & 0.2275 & \ImpP{+94.55\%} & 11.39 & \ImpP{+21.77\%} & 931.65 & \ImpP{+46.18\%} & 7959232.42 & \ImpP{+91.02\%} & 124361.14 & \ImpP{+87.70\%} \\
      TTS-GAN      & 72175.13 & \ImpP{+92.93\%} & 0.0467 & \ImpP{+73.45\%} & 13.75 & \ImpP{+35.20\%} & 958.27 & \ImpP{+47.68\%} & 12345777.01 & \ImpP{+94.21\%} & 209373.95 & \ImpP{+92.70\%} \\
      Diffusion-TS & 6741.95 & \ImpP{+24.34\%} & 0.0793 & \ImpP{+84.36\%} & 131.94 & \ImpP{+93.25\%} & 61134.11 & \ImpP{+99.18\%} & \B{609128.89} & \ImpN{-17.31\%} & \B{14966.17} & \ImpN{-2.17\%} \\
      PaD-TS       & 1738337.18 & \ImpP{+99.71\%} & 0.0969 & \ImpP{+87.20\%} & 17.06 & \ImpP{+47.77\%} & 1030.76 & \ImpP{+51.36\%} & 333745186.82 & \ImpP{+99.79\%} & 5214757.91 & \ImpP{+99.71\%} \\
      ImagenTime   & 68893.24 & \ImpP{+92.60\%} & 0.0369 & \ImpP{+66.40\%} & 12.29 & \ImpP{+27.50\%} & 943.52 & \ImpP{+46.86\%} & 13201086.53 & \ImpP{+94.59\%} & 206266.57 & \ImpP{+92.59\%} \\
      TimeAutoDiff & 6305.68 & \ImpP{+19.11\%} & \U{0.0266} & \ImpP{+53.38\%} & 13.81 & \ImpP{+35.48\%} & 960.89 & \ImpP{+47.82\%} & 1210691.10 & \ImpP{+40.98\%} & 18915.94 & \ImpP{+19.16\%} \\
      ZITS-GAN     & \U{5889.71} & \ImpP{+13.40\%} & 0.0629 & \ImpP{+80.29\%} & 12.95 & \ImpP{+31.20\%} & 948.45 & \ImpP{+47.14\%} & 1076571.42 & \ImpP{+33.63\%} & 16820.85 & \ImpP{+9.09\%} \\
      ZITS-VAE     & 5973.46 & \ImpP{+14.61\%} & 0.0463 & \ImpP{+73.22\%} & \U{9.21} & \ImpP{+3.26\%} & \U{567.74} & \ImpP{+11.69\%} & 1146881.90 & \ImpP{+37.70\%} & 17919.75 & \ImpP{+14.67\%} \\
      \midrule
      MIDiff       & \B{5100.78} & --- & \B{0.0124} & --- & \B{8.91} & --- & \B{501.37} & --- & \U{714539.49} & --- & \U{15291.30} & --- \\
      \bottomrule
    \end{NiceTabular*}
    }
  \end{threeparttable}
  \vspace{-0.5em}
\end{table*}

\subsection{Evaluation on Generation Authenticity}

\subsubsection{Metrics}
To provide a robust assessment of generation authenticity, we evaluate the synthetic data from several perspectives. The specific metrics are as follows.

\textbf{ Discriminative Accuracy~\cite{TimeGANStock}.}
This metric assesses the individual-level authenticity. It is based on a post-hoc classifier (clf) trained to distinguish between real samples (label 1) and synthetic samples (label 0). The DA score is calculated on a test set of size $S$. The goal is to achieve a score close to 0,
\begin{equation}
\text{DA} = \left| \frac{\sum_{n=1}^{S}(0=\text{clf}(\hat{x}_{n}))+\sum_{n=1}^{S}(1=\text{clf}(x_{n}))}{2S} - 0.5 \right|
\end{equation}
where $\hat{x}_{n}$ are synthetic test samples and $x_{n}$ are real test samples. Due to its instability, the mean and standard deviation are reported over 5 iterations.

\textbf{Value Distribution Shift (VDS).~\cite{PaD}}
This metric evaluates the preservation of value distributions. It is defined as the average distribution Jensen--Shannon divergence $\text{JSD}$ across all dimensions,
\begin{equation}
\text{VDS} = \frac{1}{F} \sum_{i=1}^{F} \text{JSD}(P_{V}^{i}, Q_{V}^{i})
\end{equation}
where $P_{V}^{i}$ is the distribution of the $i$-th dimension in the original \textbf{App Usage Dataset}, and $Q_{V}^{i}$ is its counterpart in the synthetic data.

\textbf{Functional Dependency Distribution Shift (FDDS).~\cite{PaD}}
This metric measures how well synthetic data preserves cross-variates dependency structures. For each unordered possible pair of dimensions $(i,j)$, we compute the Pearson correlation $\text{PC}$ in each sample between the two temporal trajectories, yielding a distribution of dependency coefficients for the original \textbf{App Usage Dataset}, $P_{FD}^{i,j}$, and for the synthetic data, $Q_{FD}^{i,j}$. FDDS is then defined as the average $\text{JSD}$ across all $M$ dimension pairs:
\begin{equation}
\text{FDDS} = \frac{1}{M} \sum_{(i,j)} \text{JSD}\left(P_{PC}^{i,j}, Q_{PC}^{i,j}\right),
\end{equation}

\begin{figure*}[!t]
    \centering
    \begin{minipage}[t]{0.49\textwidth}
        \centering
        \includegraphics[width=\linewidth]{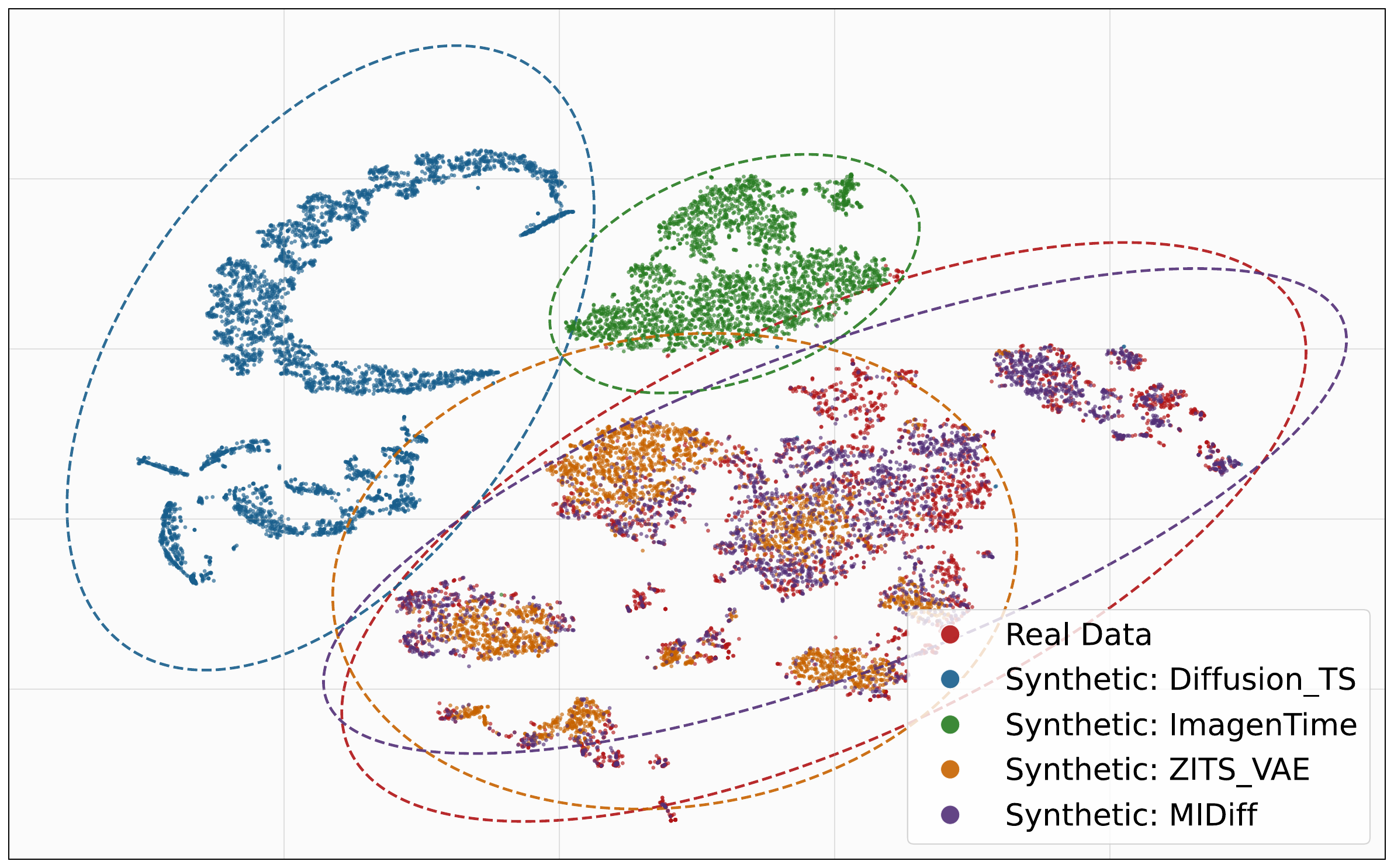}\\[-0.2em]
        \textbf{(a)}
    \end{minipage}
    \hfill
    \begin{minipage}[t]{0.49\textwidth}
        \centering
        \includegraphics[width=\linewidth]{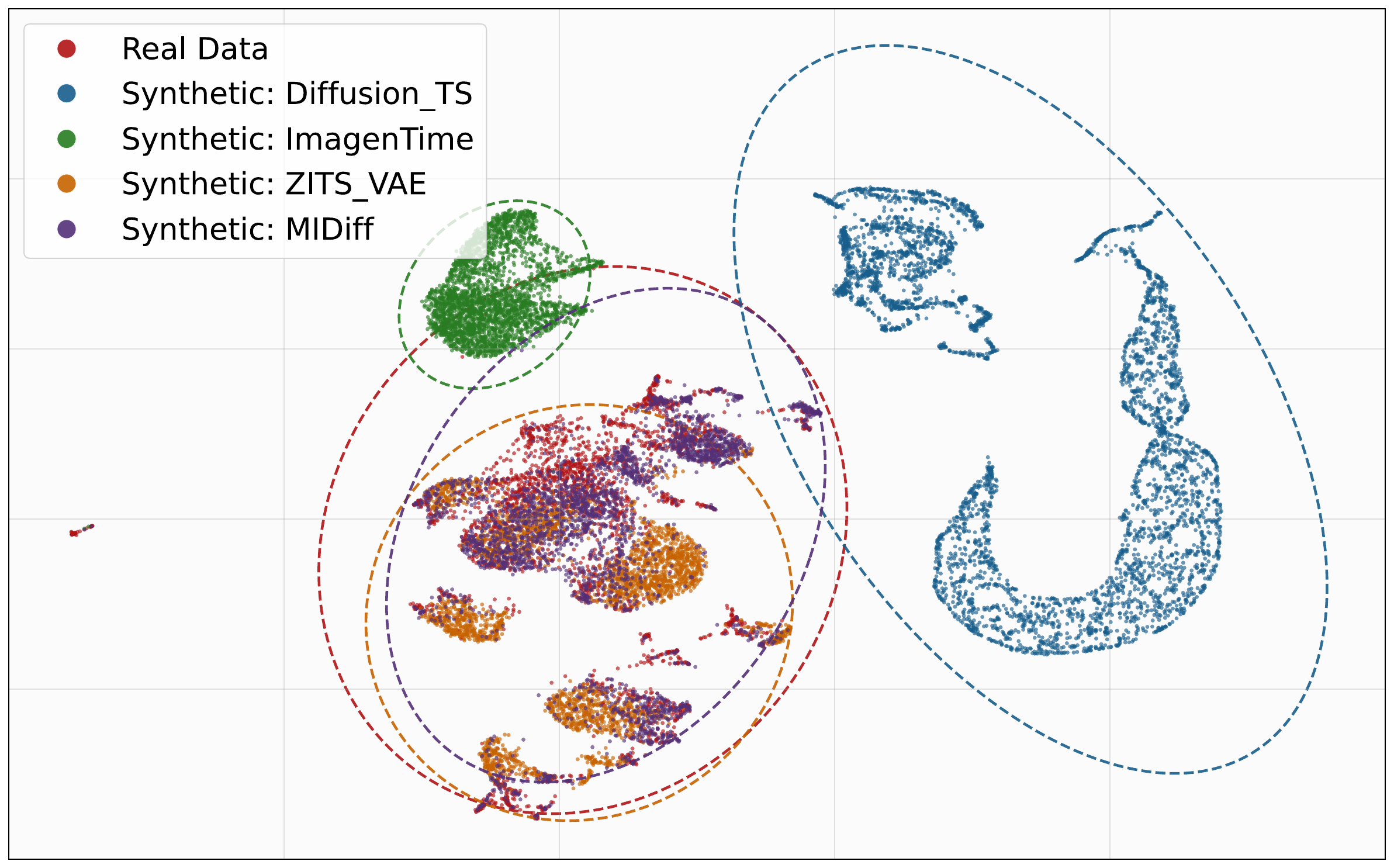}\\[-0.2em]
        \textbf{(b)}
    \end{minipage}
    \caption{Manifold comparison of generated datasets: (a) t-SNE visualization and (b) UMAP visualization.}
    \label{fig:manifold}
    \vspace{-0.5em}
\end{figure*}

 \textbf{Predictive Score~\cite{TimeGANStock}.}
This metric evaluates the utility of the generated data for downstream forecasting tasks. A post-hoc RNN model is trained on the synthetic data and then evaluated on the real data. The score itself is the resulting Mean Absolute Error (MAE) of the prediction. The mean and standard deviation are reported over 5 iterations.

\subsubsection{Results}
The quantitative results in Table~\ref{tab:scorew} demonstrate the performance of MIDiff on \textbf{App Usage Dataset}. MIDiff achieves the best results on three out of four metrics, including VDS, FDDS, and DA. For VDS, ZITS-GAN and ZITS-VAE obtain competitive results, which is consistent with their ability to model zero-inflated sparsity, since the zero/non-zero ratio is an important component of the marginal value distribution. Nevertheless, they substantially lag behind MIDiff, indicating that modeling sparsity alone is insufficient to preserve the value distribution of mobile usage traces. For FDDS, Diffusion-TS achieves the closest performance to MIDiff, suggesting its strength in capturing cross-variate dependency structures. However, its weaker results on VDS and DA show that dependency modeling alone cannot ensure faithful synthesis under sparse and heterogeneous mobile usage patterns. For DA, MIDiff shows an advantage, while most baselines obtain scores close to the upper bound of 0.5. This indicates that their generated samples can be distinguished from real traces by the post-hoc classifier, reflecting poor sample-level authenticity and inadequate fitting to the real data distribution.
\begin{table*}[!t]
\centering
\caption{Mobile usage frequency in \textbf{App Usage Dataset} (Real Data) and generated data (Timesteps = 192).}
\label{tab:usage frequency} % 新标签
% --- 表格第一部分 ---
\begin{tabular*}{\textwidth }{@{\extracolsep{\fill}}l|c|cccccccc|c@{\extracolsep{0pt}}} 
\toprule
& Real Data & TimeGAN &  TTS-GAN & Diffusion-TS & PaD-TS & ImagenTime & TimeAutoDiff & ZITS-GAN &ZITS-VAE & MIDiff \\
\midrule
Mean & 32.96  & 192  & 100.22 & 191.37  & 192 & 91.45 & 192 & 42.18 &26.79 & 37.77\\
Std & 29.40 & 0 &  4.89 & 2.40 & 0 & 15.39 & 0 & 5.81 & 8.48 & 28.16\\
\bottomrule
\end{tabular*}
\end{table*}
  \begin{figure*}[!t] % [!ht] 是一个可选参数，让LaTeX尽量把图片放在这里 (Here) 或顶部 (Top)
  \centering
\includegraphics[width=\textwidth]{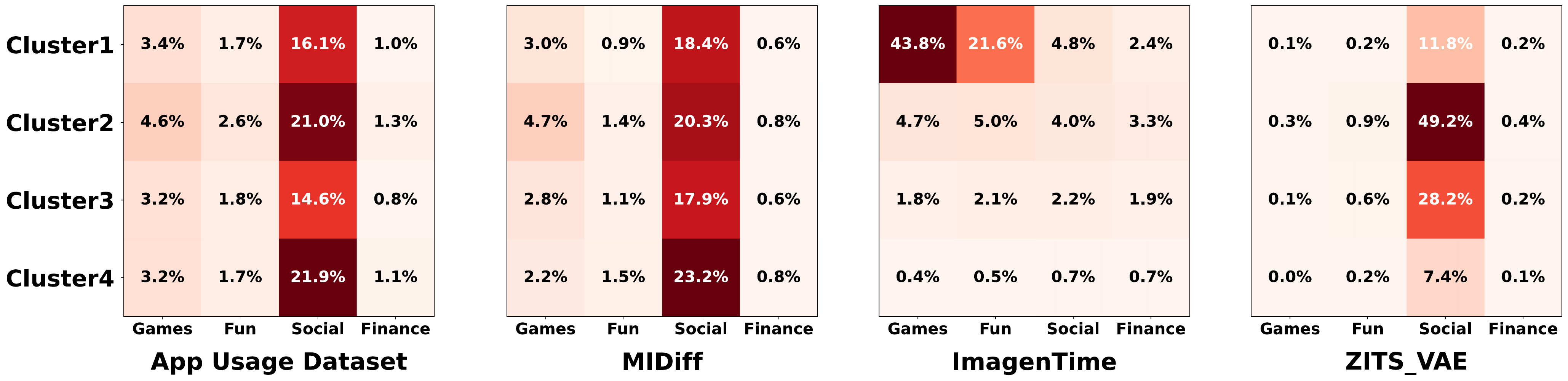}

  \caption{The heatmap illustrates different used apps in clusters of locations in different datasets.}
  \label{fig:correlation-heatmap}
\end{figure*}
  \begin{figure*}[!t] % [!ht] 是一个可选参数，让LaTeX尽量把图片放在这里 (Here) 或顶部 (Top)
  \centering
\includegraphics[width=\textwidth]{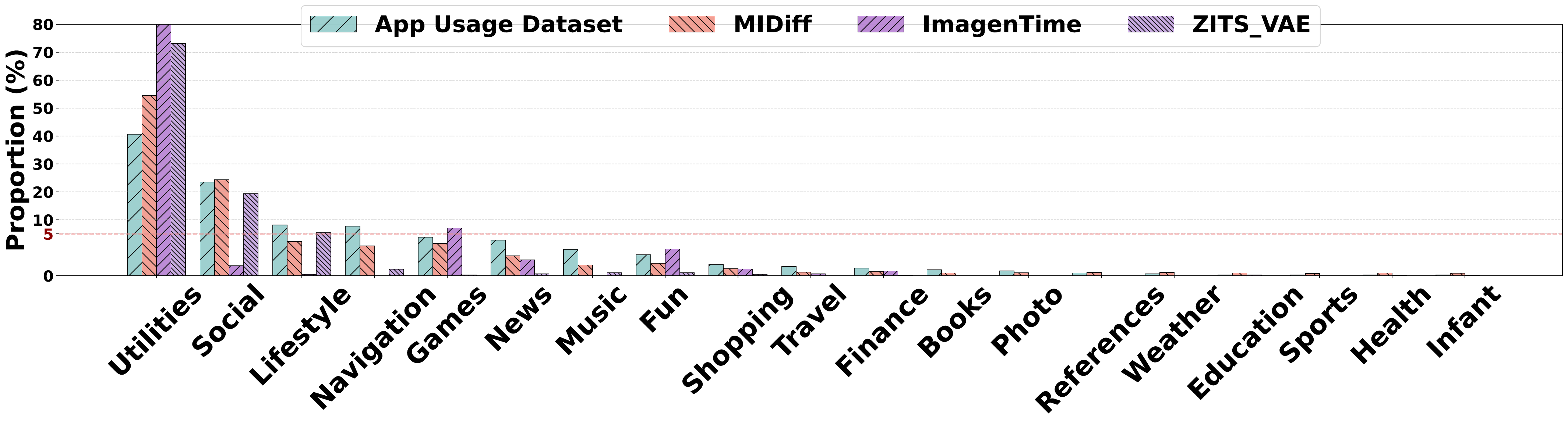}
   \caption{A comparison of the app usage distribution between \textbf{App Usage Dataset} and generated data, ranked by the app usage frequency in \textbf{App Usage Dataset}. }
   \label{fig:App distribution} % 整个图的引用标签
 \end{figure*}
The only metric where MIDiff does not rank first is Predictive Score. ZITS-VAE, ZITS-GAN, and TimeAutoDiff obtain slightly lower scores, while MIDiff remains competitive with a score of 7296. However, their weaker performance on VDS, FDDS, and DA indicates that better predictive utility does not necessarily correspond to better global distributional alignment. Overall, MIDiff achieves clear advantages in distributional fidelity. This conclusion is reinforced by the feature and distance-based measures on TS-bench~\cite{TSGBENCH} presented in Table \ref{tab:results_table3_all}.

\subsubsection{Manifold Analysis}
To qualitatively evaluate the distributional fidelity of the generated mobile usage traces, we project the high-dimensional representations of real and synthetic samples into two-dimensional spaces using t-SNE and UMAP. As shown in Figure~\ref{fig:manifold}, MIDiff exhibits the closest manifold alignment with \textbf{App Usage Dataset} across both projections, with substantial overlap in the central region and better coverage of peripheral clusters.

In contrast, Diffusion-TS and ImagenTime show clear distributional shifts, as their generated samples form manifolds that are largely separated from \textbf{App Usage Dataset}. ZITS achieves better alignment around the high-density central region, but its samples cover only a subset of the real manifold and miss several peripheral regions. This suggests insufficient support coverage, especially for less frequent usage patterns. MIDiff more closely follows the shape of the real manifold, including both the dense central area and the scattered peripheral regions.

\begin{table*}[!t]
\caption{
Downstream performance on the cross-variate prediction for the model trained on \textbf{App Usage Dataset} with different augment data from different generative models. 
The final $\Delta$ row shows gain or drop against the model trained only by \textbf{App Usage Dataset}.
}
\renewcommand{\arraystretch}{0.85}

% Requires:
%\usepackage{booktabs}
% \usepackage{multirow}
%\usepackage[table]{xcolor}
%\usepackage{nicematrix}

\definecolor{posGreen}{RGB}{50,130,90}
\definecolor{topGray}{RGB}{242,242,242}
\definecolor{midiffBlue}{RGB}{235,245,255}

\newcommand\Sc[1]{{\footnotesize #1}}
\newcommand\ImpP[1]{\textcolor{posGreen}{\scriptsize #1}}
\newcommand\ImpN[1]{\textcolor{gray}{\scriptsize #1}}
\newcommand\ImpPD[1]{\textcolor{posGreen}{\footnotesize#1}}
\newcommand\ImpND[1]{\textcolor{gray}{\footnotesize #1}}
\newcommand\Best[1]{\textbf{\footnotesize #1}}
\newcommand\Second[1]{\underline{\footnotesize #1}}
\newcommand\G[1]{\textcolor{gray}{\footnotesize #1}}
\newcommand\Head[1]{{\small\textbf{#1}}}
\newcommand\RowName[1]{{\small #1}}
\newcommand\ModelRow[1]{\multirow{2}{*}{\RowName{#1}}}

\centering
\resizebox{\textwidth}{!}{%
    \setlength{\tabcolsep}{1.5pt}
    % Make rule-adjacent background filling cleaner.
    \setlength{\aboverulesep}{0pt}
    \setlength{\belowrulesep}{0pt}
    \begin{NiceTabular}{l*{12}{c}}
    \CodeBefore
        % Fill the top header block, bounded by \toprule and \midrule.
        \rectanglecolor{topGray}{1-1}{3-13}
        % Fill the MIDiff block, bounded by \midrule and the following Delta row.
        \rectanglecolor{midiffBlue}{9-1}{11-13}
    \Body
    \toprule
    \addlinespace[0.5em]
    \Head{Target}
    & \multicolumn{6}{c}{\Head{Traffic}} 
    & \multicolumn{3}{c}{\Head{Application}} 
    & \multicolumn{3}{c}{\Head{Location Cluster}} \\
    \cmidrule(lr){2-7} \cmidrule(lr){8-10} \cmidrule(l){11-13}
    \addlinespace[0.3em]
    \Head{Model}
    & \multicolumn{2}{c}{\Head{LSTM}} 
    & \multicolumn{2}{c}{\Head{MLP}} 
    & \multicolumn{2}{c}{\Head{Mamba}} 
    & \Head{LSTM}
    & \Head{MLP}
    & \Head{Mamba}
    & \Head{LSTM}
    & \Head{MLP}
    & \Head{Mamba} \\
    \cmidrule(lr){2-3} \cmidrule(lr){4-5} \cmidrule(lr){6-7}
    \cmidrule(lr){8-8} \cmidrule(lr){9-9} \cmidrule(lr){10-10}
    \cmidrule(lr){11-11} \cmidrule(lr){12-12} \cmidrule(lr){13-13}
    \addlinespace[0.3em]
    \Head{Metric}
    & \Head{MSE} & \Head{$R^2$} 
    & \Head{MSE} & \Head{$R^2$} 
    & \Head{MSE} & \Head{$R^2$} 
    & \Head{Macro-F1} 
    & \Head{Macro-F1} 
    & \Head{Macro-F1} 
    & \Head{Macro-F1} 
    & \Head{Macro-F1} 
    & \Head{Macro-F1} \\
    \midrule
    \addlinespace[0.3em]
    \RowName{\textbf{App Usage Dataset}}
    & \Second{2.04e11} & \Second{0.0101}
    & \Second{2.05e11} & \Second{0.0040}
    & \Sc{2.10e11} & \Sc{-0.0207}
    & \Second{0.0467}
    & \Best{0.0867}
    & \Sc{0.0485}
    & \Sc{0.2068}
    & \Second{0.3005}
    & \Second{0.1042} \\
    \addlinespace[0.5em]

    \ModelRow{+ ImagenTime's}
    & \Sc{2.83e11} & \Sc{-0.3740}
    & \Sc{2.72e11} & \Sc{-0.3209}
    & \Sc{3.77e11} & \Sc{-0.8285}
    & \Best{0.0536}
    & \Sc{0.0488}
    & \Best{0.0653}
    & \Sc{0.1071}
    & \Sc{0.0683}
    & \Sc{0.0986} \\[-0.05em]
    & \ImpN{-38.73\%} & \ImpN{-0.3841}
    & \ImpN{-32.68\%} & \ImpN{-0.3249}
    & \ImpN{-79.52\%} & \ImpN{-0.8078}
    & \ImpP{+0.69\%}
    & \ImpN{-3.79\%}
    & \ImpP{+1.68\%}
    & \ImpN{-9.97\%}
    & \ImpN{-23.22\%}
    & \ImpN{-0.56\%} \\
    \addlinespace[0.3em]

    \ModelRow{+ ZITS\_VAE's}
    & \Sc{2.08e11} & \Sc{-0.0090}
    & \Sc{2.06e11} & \Sc{-0.0006}
    & \Second{2.04e11} & \Second{0.0094}
    & \Sc{0.0448}
    & \Sc{0.0618}
    & \Second{0.0580}
    & \Second{0.2256}
    & \Sc{0.2918}
    & \Sc{0.0940} \\[-0.05em]
    & \ImpN{-1.96\%} & \ImpN{-0.019}
    & \ImpN{-0.49\%} & \ImpN{-0.005}
    & \ImpP{+2.86\%} & \ImpP{+0.030}
    & \ImpN{-0.19\%}
    & \ImpN{-2.49\%}
    & \ImpP{+0.95\%}
    & \ImpP{+1.88\%}
    & \ImpN{-0.87\%}
    & \ImpN{-1.02\%} \\

    \midrule
        \addlinespace[0.3em]
    \ModelRow{+ MIDiff's}
    & \Best{2.02e11} & \Best{0.0203}
    & \Best{1.97e11} & \Best{0.0438}
    & \Best{1.97e11} & \Best{0.0429}
    & \Sc{0.0409}
    & \Second{0.0626}
    & \Sc{0.0560}
    & \Best{0.3054}
    & \Best{0.3179}
    & \Best{0.1180} \\[-0.05em]
    & \ImpP{+0.98\%} & \ImpP{+0.010}
    & \ImpP{+3.90\%} & \ImpP{+0.040}
    & \ImpP{+6.19\%} & \ImpP{+0.064}
    & \ImpN{-0.58\%}
    & \ImpN{-2.41\%}
    & \ImpP{+0.75\%}
    & \ImpP{+9.86\%}
    & \ImpP{+1.74\%}
    & \ImpP{+1.38\%}\\
    $\Delta$ 
    & \ImpPD{+0.98\%} & \ImpPD{+0.010}
    & \ImpPD{+3.90\%} & \ImpPD{+0.040}
    & \ImpPD{+3.43\%} & \ImpPD{+0.034}
    & \ImpND{-1.27\%}
    & \ImpND{-2.41\%}
    & \ImpND{-0.93\%}
    & \ImpPD{+7.98\%}
    & \ImpPD{+1.74\%}
    & \ImpPD{+1.38\%} \\

    \bottomrule
    \end{NiceTabular}%
}
\label{Downstream}

\end{table*}
 
\subsection{Evaluation on Mobile Usage Trace Features}

We further evaluate the models' ability to capture the three unique characteristics of mobile usage traces previously identified in the Introduction, (1) sparse user behavior, (2) strong inter-variable correlations, and (3) unbalanced app usage frequencies. For visualization, we select ImagenTime and TTS-GAN for comparison, as they exhibit the highest visual fidelity among the baselines.

\textbf{Challenge 1: Data Sparsity Modeling.}
To evaluate the modeling of behavioral sparsity, we compare the mobile usage frequency of generated traces with that of real users. As shown in Table~\ref{tab:usage frequency}, MIDiff achieves the closest mean usage frequency to \textbf{App Usage Dataset}, with 37.77 compared to the real value of 32.96. Although the ZITS variants also obtain relatively close mean values, their standard deviation remain much smaller than the real distribution. Similar collapse is observed in other baselines. In contrast, MIDiff achieves a standard deviation of 28.16, closely matching \textbf{App Usage Dataset}. These results indicate that MIDiff not only captures the mean sparsity level of mobile usage behaviors, but also preserves the diversity of usage frequencies across users, demonstrating better alignment with real usage patterns.

\textbf{Challenge 2: Correlation Modeling.}
To analyze inter-variable correlation and unbalanced usage, we visualize the proportion of usage for specific app categories within different location clusters. The results are presented as a heatmap in Figure~\ref{fig:correlation-heatmap}. \textbf{App Usage Dataset} contains a complex and unbalanced pattern, ``Social'' apps are dominant in these locations. App usage in these 4 types of apps is strongly correlated with location, appearing significantly more in locations belonging to clusters 2 and 4 than in clusters 1 and 3. ``Games'' are used sparingly, while ``Fun'' and ``Finance'' are almost unused. MIDiff is the only model that successfully captures this correlation. In contrast, ImagenTime fails to generate any meaningful app usage for locations not belonging to clusters 1. ZITS correctly captures the high frequency of ``Social'' but fails on the correlation, incorrectly attributing the main usage to locations in cluster 2 and generating excessive usage for the ``Fun''.

\textbf{Challenge 3: Imbalanced App Usage Modeling.}
We analyze the model's ability to capture the highly unbalanced app usage frequencies. We compute the usage proportion for each app category across both \textbf{App Usage Dataset} and generated data, as shown in Figure~\ref{fig:App distribution}. \textbf{App Usage Dataset} exhibits a highly imbalanced distribution, ``Utilities'' is the most frequently used category, followed by ``Social''. Most other categories are used much more sparingly. MIDiff, which utilizes different convolutional kernels to independently learn the distribution for each app in C-GASF, successfully replicates this distribution. It correctly identifies ``Utilities'' as the primary app category while also generating a substantial number of ``Social'' usages. Crucially, for rarely used categories like ``Navigation'' and ``Lifestyle'', MIDiff avoids both over-generation and complete mode collapse. In contrast, baselines like ImagenTime and ZITS are negatively impacted by the usage imbalance. ImagenTime overproduces the dominant ``Utilities'' category and poorly models secondary usage categories such as ``Social'' and ``Lifestyle.'' ZITS, in contrast, captures the three most frequent categories reasonably well, but fails to follow the real distribution for medium-frequency categories such as ``Navigation,'' ``News,'' and ``Music.'' These results indicate that both models exhibit biased coverage over the long-tailed category distribution, thereby reducing the authenticity and diversity of the generated mobile usage traces.

To sum up, while none of the baselines successfully modeled any unique mobile usage trace characteristics, MIDiff successfully modeled all of them, generating data that authentically conforms to the real app usage distribution.

\begin{table*}[!t]
\centering
\caption{Ablation study of generative backbones and attention modules.}
\label{tab:scorew_ablated}
\begin{threeparttable}
\small
\setlength{\tabcolsep}{4pt}
\renewcommand{\arraystretch}{0.65}
\setlength{\aboverulesep}{0.3ex}
\setlength{\belowrulesep}{0.3ex}
\setlength{\cmidrulesep}{0.2ex}

\begin{tabular*}{\textwidth}{@{\extracolsep{\fill}} l cc cccc @{}}
\toprule
\multirow{2}{*}{\textbf{Method}} 
& \multicolumn{2}{c}{\textbf{Attention Module}} 
& \multicolumn{4}{c}{\textbf{Metrics} ($\downarrow$)} \\
\cmidrule(lr){2-3} \cmidrule(lr){4-7}
& \textbf{Basic} & \textbf{Triplet} 
& \textbf{VDS} & \textbf{FDDS} 
& \textbf{DA $[0,0.5]$} & \textbf{Pred. Score} \\
\midrule
GAN~\cite{gan}
& \multicolumn{2}{c}{All Variants} 
& \multicolumn{4}{c}{\textit{Failed to converge}} \\
\midrule
\multirow{3}{*}{VAE~\cite{vae}} 
& \checkmark & - 
& \multicolumn{4}{c}{\multirow{2}{*}{\textit{Failed to form C-GASF patterns}}} \\
& \checkmark & \checkmark 
& \multicolumn{4}{c}{} \\
& - & \checkmark 
& 0.023 & 0.214 & $0.421\pm0.002$ & $6564\pm8$ \\
\midrule
\multirow{3}{*}{\textbf{Diffusion}} 
& \checkmark & - 
& 0.021 & 0.255 & $0.422\pm0.005$ & $6421\pm9$ \\
& \checkmark & \checkmark 
& 0.015 & 0.148 & $0.448\pm0.007$ & $6619\pm40$ \\
& - & \checkmark 
& \textbf{0.002} & \textbf{0.029} 
& $\mathbf{0.153\pm0.006}$ & $\mathbf{7296\pm85}$ \\
\bottomrule
\end{tabular*}
\end{threeparttable}
\vspace{-0.5em}
\end{table*}

\subsection{Evaluation on Downstream Utility via Synthetic Data Augmentation}

We evaluated the practical utility of the generated traces through an augmentation-based downstream evaluation in multivariate-to-univariate forecasting~\cite{timexer,tqnet}. The original traces were represented as multivariate sequences of length 192 with three channels, corresponding to traffic volume, App, and Location. For each sample, we randomly selected a target timestamp from the second half of the sequence where the traffic channel was non-zero, and used the preceding 96 timesteps of the other two variables as input to predict the held-out target variable. This yielded three cross-variable downstream tasks: predicting Traffic from App and Location, predicting App from Traffic and Location, and predicting Location from Traffic and App. We evaluated three backbone models, including MLP, LSTM, and Mamba~\cite{mamba}. For the discrete targets App and Location, we formulated the task as classification and reported Macro-F1. For the continuous Traffic target, we formulated the task as regression and reported both raw-scale MSE and $R^2$. 
The raw-scale MSE prevents the evaluation from being overly diluted by normalized scales or by predictions biased toward near-zero values and $R^2$ further assesses whether the generated data preserves meaningful cross-variable predictive dependencies rather than only matching average traffic levels. Table~\ref{Downstream} summarizes the downstream performance of MIDiff-generated augmentation against the competing generative baselines.

The results in Table~\ref{Downstream} show that existing generative baselines fail to provide consistent augmentation benefits: none of them improves all backbone models for any target variable. In contrast, augmenting the real training set with MIDiff-generated traces consistently improves Traffic and Location prediction, while also achieving the best results for these two targets. For Traffic, although training on \textbf{App Usage Dataset} alone already yields positive $R^2$ values for LSTM and MLP, augmenting with ImagenTime or ZITS-VAE makes the $R^2$ scores negative under both backbones. MIDiff, instead, increases the $R^2$ from 0.0101 to 0.0203 for LSTM and from 0.0040 to 0.0438 for MLP, more than doubling the explained variance. It also turns the Mamba $R^2$ from -0.0207 to 0.0429, indicating that MIDiff better preserves the cross-variate dependencies from App and Location to Traffic. For Location, MIDiff also achieves the best performance across all backbones, showing that the generated traces effectively improve behavior modeling.

For Application, MIDiff does not obtain the best score under any single backbone, but no generative baseline consistently outperforms it across all backbones. With MLP, MIDiff achieves the strongest generative-augmentation result, reaching 0.0626, slightly higher than ZITS-VAE with 0.0618 and ImagenTime with 0.0488, although still lower than training on \textbf{App Usage Dataset}. With Mamba, MIDiff improves over the \textbf{App Usage Dataset} only, increasing Macro-F1 from 0.0485 to 0.0560. Overall, MIDiff-generated traces provide the most reliable augmentation, improving Traffic and Location prediction while maintaining competitive performance for Application prediction.

 \subsection{Ablation Study}
The ablation study investigates the capability of different generative methods and attention block configurations in synthesizing authentic C-GASF patterns, with results summarized in Table \ref{tab:scorew_ablated}. The GAN-based models consistently fail to yield results due to mode collapse. The VAE-based models struggle with capturing the fine-grained features necessary for C-GASF generation and inverse transformation. They yield competitive results only when the Triplet Attention module is used exclusively, highlighting its necessity. In contrast, the Diffusion models demonstrate superior training stability and robust fine-grained feature recovery, successfully generating C-GASF patterns across all attention configurations. The results confirm the critical role of the Triplet Attention module in representation learning. The Diffusion model achieves its best performance in DA $0.153$ when using this module alone. The significant difference in DA scores between the Origin-based Diffusion model score $0.422$ and the Triplet-based Diffusion model underscores the incompatibility of the Original Attention block with C-GASF patterns and the superior adaptability of the Triplet Attention mechanism. Finally, all diffusion-based models successfully generating the C-GASF pattern achieve markedly lower DA scores compared to the previous baseline VRAE, whose $\text{DA}=0.489$. Overall, these results demonstrate that the combination of Diffusion modeling and the Triplet Attention module effectively captures the intrinsic structure of the C-GASF image space, enabling stable training and accurate fine-grained representation learning.

\section{Conclusion}
In this study, we introduced MIDiff, a diffusion-based framework for generating realistic user-level mobile traces. To better represent sparse and heterogeneous mobile behaviors, we proposed C-GASF, which transforms multivariate usage traces into structured image patterns and helps preserve three key characteristics: sparse usage, heterogeneous cross-channel dependencies, and long-tail application usage patterns. We further incorporated Triple Attention to the U-Net model to capture temporal dynamics and cross-variable consistency. Extensive experiments show that MIDiff consistently outperforms existing time-series generation baselines and better reconstructs the characteristics of real-world data.

Future work will explore conditional generation under different user preferences and examine the generalizability of C-GASF across diverse datasets.

\ifCLASSOPTIONcaptionsoff
  \newpage
\fi



%\bibliographystyle{IEEEtran}
\bibliographystyle{IEEEtran}  % 选择格式：IEEEtran, plain, unsrt, alpha等
\bibliography{references} 



% \end{thebibliography}


% \end{IEEEbiography}


% % insert where needed to balance the two columns on the last page with
% %\newpage



%\vfill

\newpage
\appendices
\section{Inverse Transform of Cross-Gramian Angular Sum Field}
\vspace{-2mm} 
\begingroup
\small % Use smaller font for these two algorithms
\begin{algorithm}[H]
\caption{Single Image Decoding}
\label{alg:process_image}
\begin{algorithmic}[1]
\Require Image $\mathbf{I} \in \mathbb{R}^{1\times T \times C(C'+1)}$, timesteps $T$, the number of PoI clusters $C'$ and app categories $C$. $
l_{\max,i}
=
\max_{u \in U,\ \tau \in \{1,\dots,T\}}
A_{\tau,i}^{(u)}$, $\bm{l}_{\max} \in \mathbb{R}^{C}$denoted per-app traffic usage maximum

\Ensure Decoded $(\mathbf{V,A}, \mathbf{P'})$

\Procedure{ProcessImage}{$\mathbf{I}, \bm{l}_{\max}, C, C'$}
     \State $\mathbf{Av} \gets \mathbf{0}_{T \times C}$, $\mathbf{P}^{\text{init}} \gets \mathbf{0}_{T \times (C'+1)}$ \Comment{Initialization}
    \State $\mathbf{G} \gets \text{Reshape}(\mathbf{I}, (T, C, C'+1))$
    
    \ForAll{$t = 1 \to T$}  
\State $v_i \gets \arg\max_{j \in \{0,\dots,C'\}} G_{t,i,j}, \quad i=1,\dots,C$
\Comment{$\bm{v} \in \{0,\dots,C'\}^{C}$}
        \State $p_{\text{mode}} \gets \text{Mode}(\bm{v})$
        
        \If{$p_{\text{mode}} = 0$} \Comment{No usage}
            \State $\mathbf{P^{\text{init}}}_{t, 0} \gets 1$
            \State \textbf{continue} \Comment{Leave $\mathbf{A}_t$ as $\mathbf{0}$}
        \Else \Comment{Find app block}
            \State $\mathbf{P^{\text{init}}}_{t, p_{\text{mode}}} \gets 1$
            \State $\bm{j}^{(2)} \gets \text{ArgSort}(\mathbf{{G}_t}, \text{axis}=1)[:,-2]$ \Comment{Indices of 2nd largest values}
            \State $\bm{r} \gets \mathbf{0}_{C}$
            
            \ForAll{$i = 1 \to C$}
                \State $\bm{M}^{(i)} \gets \mathbf{1}_{C'+1}$
                \State $M^{(i)}_{p_{\text{mode}}} \gets 0, \quad M^{(i)}_{j^{(2)}_i} \gets 0$
                \Comment{Exclude the dominant PoI column and row-wise second-largest response.}
                \State {\scriptsize$r_i \gets 
                \begin{cases}
                \dfrac{\sum_{j=0}^{K-1} {G}_{t,i,j} M^{(i)}_j}{\sum_{j=0}^{K-1} M^{(i)}_j}, 
                & \text{if } \sum_{j=0}^{K-1} M^{(i)}_j > 0,\\
                -1, & \text{otherwise}.
                \end{cases}$}
            \EndFor
            
            \State $a_{\text{pos}} \gets \arg\max_i \bm{r}$ \Comment{Largest average response}
            \State $\mathbf{A^{\text{init}}}_{t, a_{\text{pos}}} \gets \mathbf{{G}_{t,a_\text{pos}, p_{\text{mode}}}}$
        \EndIf
    \EndFor
    
    \State $\mathbf{A'} \gets \text{Inverse-Normalization (Alg. 2)}(\mathbf{A^{\text{init}}}, \bm{l}_{\max})$ 
    \State $\mathbf{V},\mathbf{A}\gets \text{One-hotDecoding($A'$)}$
    \State \Return $(\mathbf{V}, \mathbf{A},\mathbf{P'})$
\EndProcedure
\end{algorithmic}
\end{algorithm} 

\begin{algorithm}[H]
\caption{Inverse-Normalization}
\begin{algorithmic}[1]
\Require Normalized trace $\mathbf{A^{\text{init}}} \in \mathbb{R}^{T \times C}$, maxima $\bm{l}_{\max} \in \mathbb{R}^{C}$
\Ensure Inverse-normalized trace $\mathbf{A'} \in \mathbb{R}^{T \times C}$

\Procedure{InverseTransform}{$\mathbf{A^{\text{init}}}, \bm{l}_{\max}$}
    \If{$\mathbf{A^{\text{init}}} = \mathbf{0}$} \Comment{Filter 1: Check for all-zero input}
        \State \Return $\mathbf{A^{\text{init}}}$
    \EndIf

    \State $\mathbf{A'} \gets \mathbf{A^{\text{init}}} \odot \bm{l}_{\max}$ \Comment{Element-wise broadcasted multiplication}
    \State \Return $\mathbf{A'}$
\EndProcedure
\end{algorithmic}
\end{algorithm}
\label{alg:inverse_transform}

\endgroup
\end{document}